\documentclass[conference]{IEEEtran}
\makeatletter
\def\ps@IEEEtitlepagestyle{%
  \def\@oddfoot{\mycopyrightnotice}%
  \def\@evenfoot{}%
}

\usepackage{blindtext}
\usepackage{eso-pic}
\IEEEoverridecommandlockouts
\usepackage{cite}
\usepackage{amsmath,amssymb,amsfonts}
\usepackage{algorithmic}
\usepackage{graphicx}
\usepackage{textcomp}
\usepackage{xcolor}
\usepackage{comment}
\usepackage{booktabs}
\usepackage{bm}
\usepackage{subcaption}
\usepackage{tikz}
\usepackage{circuitikz}
\usepackage{xurl}
\usepackage[hidelinks]{hyperref}
\usepackage[table]{xcolor}
\usetikzlibrary{arrows.meta}
\def\BibTeX{{\rm B\kern-.05em{\sc i\kern-.025em b}\kern-.08em
    T\kern-.1667em\lower.7ex\hbox{E}\kern-.125emX}}

\usepackage{eso-pic}

\newcommand{\linebreakand}{%
  \end{@IEEEauthorhalign}
  \hfill\mbox{}\par
  \mbox{}\hfill\begin{@IEEEauthorhalign}
}
\begin{document}
\title{\vspace*{1cm} Multi-Class Electrical and Mechanical Fault Classification Using Random Convolutional Kernels\\
}
\author{\IEEEauthorblockN{Mouhamadou Mansour LO}
\IEEEauthorblockA{\textit{UR 4025, LSEE} \\
\textit{University of Artois}\\
F-62400 Béthune, France \\
mouhamadou.lo@univ-artois.fr}
\and
\IEEEauthorblockN{Mouad TALBAOUI}
\IEEEauthorblockA{\textit{UR 4025, LSEE} \\
\textit{University of Artois}\\
F-62400 Béthune, France \\
mouad.talbaoui@univ-artois.fr}
\and
\IEEEauthorblockN{Gildas MORVAN}
\IEEEauthorblockA{\textit{UR 3926, LGI2A} \\
\textit{University of Artois}\\
F-62400 Béthune, France \\
gildas.morvan@univ-artois.fr}
\linebreakand
\IEEEauthorblockN{Mathieu ROSSI}
\IEEEauthorblockA{\textit{UR 4025, LSEE} \\
\textit{University of Artois}\\
F-62400 Béthune, France \\
mathieu.rossi@univ-artois.fr}
\and
\IEEEauthorblockN{Fabrice MORGANTI}
\IEEEauthorblockA{\textit{UR 4025, LSEE} \\
\textit{University of Artois}\\
F-62400 Béthune, France \\
fabrice.morganti@univ-artois.fr}
\and
\IEEEauthorblockN{David MERCIER}
\IEEEauthorblockA{\textit{UR 3926, LGI2A} \\
\textit{University of Artois}\\
F-62400 Béthune, France \\
david.mercier@univ-artois.fr}
}
\maketitle
\begin{abstract}
Diagnosing faults in rotating machinery is essential for ensuring the reliability of industrial processes. Random convolutional kernel-based Time Series Classification (TSC) methods, such as ROCKET and its variants, provide an attractive trade-off between predictive performance and computational efficiency. In this work, we evaluate \emph{SelF-Rocket} for the multi-class diagnosis of both mechanical and electrical faults and introduce, as a new contribution, a multivariate extension of the original method. The proposed approach is compared with leading ROCKET-based methods on two public benchmark datasets, MaFaulDa (mechanical faults) and ITSC-UDG (stator inter-turn short circuits), under both univariate and multivariate settings. Experimental results show that \emph{SelF-Rocket} achieves the best overall accuracy--latency trade-off among the evaluated methods, obtaining the highest classification performance on MaFaulDa while remaining highly competitive on the more challenging ITSC-UDG dataset.
\end{abstract}
\section{Introduction}
Fault diagnosis in rotating machinery is a recurring problem in industrial processes~\cite{hamani2025}. Numerous methods have been developed to derive indicators that help anticipate the degradation of a mechanical component (bearing, imbalance, misalignment, eccentricity) or the occurrence of an electrical fault (inter-turn short-circuit). Classical methods often rely on frequency-domain analysis and require expert (physical) knowledge of these faults as well as of the machine characteristics (number of poles, rotational speed, operating point)~\cite{ceban2011,miftah2019}. They can therefore be difficult to implement, poorly generalizable, and not always conclusive in practice.
In recent years, many methods based on machine learning, and in particular deep learning, have emerged~\cite{gultekin2023}. While they achieve very good performance, they require large amounts of data and substantial computational resources.
To balance performance and computational cost, methods based on random convolutional kernels, introduced with ROCKET~\cite{dempster2020rocket}, have recently attracted considerable interest for time series classification. Rather than learning the convolution filters, these approaches generate a large number of them randomly and extract simple descriptors through a pooling operator, which are then passed to a linear classifier. They thus reach performance comparable to that of deep learning while drastically reducing training time and removing the need for any prior expert knowledge of the faults or of the machine characteristics. Many variants have since been proposed, such as
MiniRocket~\cite{dempster2021minirocket}, MultiRocket~\cite{tan2022multirocket}, and Hydra~\cite{dempster2023hydra}, improving the trade-off between accuracy and efficiency.
Most of these methods rely on one or more fixed pooling operators, such as
the Proportion of Positive Values (PPV), combined with a predefined input
representation to extract features from time series. However, although PPV
is a sound default, it is not necessarily the optimal operator for every
dataset~\cite{lo2026timeseriesclassificationrandom}.  Building on this observation, SelF-Rocket (Selected Features Rocket)~\cite{lo2026timeseriesclassificationrandom}, based on MiniRocket, dynamically selects during training the best Input Representation-Pooling Operator (IR-PO) pair from a set of candidates, rather than relying on a single combination fixed a priori.
In this article, we introduce a multivariate extension of SelF-Rocket and evaluate it for the multi-class diagnosis of both mechanical and electrical faults in rotating machines. Both the univariate and multivariate versions are compared with the main ROCKET-based approaches on two public datasets, MaFaulDa~\cite{ribeiro2018mafaulda} (mechanical faults) and ITSC-UDG~\cite{CARDENASCORNEJO2023113680} (inter-turn short-circuits), while considering both predictive and computational efficiency.
The remainder of this article is organized as follows. The two datasets are described in Section~\ref{sec:datasets}, and the proposed approach is presented in Section~\ref{sec:model}. Results are then reported and analyzed in Section~\ref{sec:results}. Finally, concluding remarks and some prospects are given in Section~\ref{sec:conclusion}.
\section{Datasets Description}\label{sec:datasets}
\subsection{MaFaulDa Dataset}
The Machinery Fault Database (MaFaulDa) is a publicly available benchmark dataset for mechanical fault diagnosis in rotating machinery. It comprises 1,951 multivariate time series, each recorded over 5 seconds at a sampling frequency of 50 kHz using eight measurement channels: three vibration channels from a triaxial accelerometer (radial, axial, and tangential), three additional single-axis vibration channels, one tachometer channel, and one microphone channel. The data were acquired using the SpectraQuest Machinery Fault Simulator (MFS) Alignment-Balance-Vibration Trainer (ABVT).
The dataset covers one healthy condition and nine fault conditions, including rotor imbalance, horizontal and vertical misalignment, and bearing defects at both the underhang and overhang positions. Bearing faults are further categorized according to the defect location (cage, outer race, and ball), resulting in a total of ten classes.
The recordings correspond to different combinations of fault type, rotational speed, and fault severity. Rotational speeds range from 737 to 3,686 rpm over approximately 49 operating points with increments of about 60 rpm, providing a wide variety of operating scenarios for fault diagnosis.
\subsection{ITSC-UDG Dataset}
The ITSC-UDG (University of Guanajuato) dataset is a publicly available benchmark dataset for electrical fault diagnosis in induction motors. It comprises 65 multivariate time series of the three simultaneously measured stator phase currents, each recorded over 5 seconds at a sampling frequency of 1 kHz. The dataset focuses on the diagnosis of inter-turn short-circuit (ITSC) faults in the stator windings of a three-phase squirrel-cage induction motor. The authors provide two versions of the dataset: a raw version containing 5,000 samples per recording and a preprocessed version containing 1,000 samples per recording, obtained by filtering the signals and retaining only the steady-state fault region. In this work, we use the raw version in order to apply the same preprocessing pipeline as that adopted for the MaFaulDa dataset.
It includes 13 operating conditions: one healthy condition and 12 ITSC fault conditions corresponding to the three stator phases and four fault severity levels (10\%, 20\%, 30\%, and 40\% of the affected winding turns). Consequently, the dataset enables the identification of both the faulty phase and the fault severity.
All measurements were acquired under steady-state no-load operating conditions at 1,725 rpm with a 60 Hz power supply. Five independent repetitions were recorded for each operating condition.
\section{Proposed Model}\label{sec:model}
\subsection{SelF-Rocket for Multivariate Time Series Classification}
\begin{figure*}[!t]
    \centering
    \resizebox{1\textwidth}{!}{%
    \begin{circuitikz}

\tikzset{
    every node/.style={font=\Large},
    flowarrow/.style={
        line width=1.4pt,
        -{Stealth[length=4mm,width=2.5mm]}
    }
}

\tikzset{
    fullarrow/.style={
        line width=3pt,
        -{Triangle[length=5mm,width=4mm]}
    }
}


\draw[domain=-7:-1.5,samples=120,smooth]
plot (\x,{15.9 + 0.35*sin(5*\x r) + 0.18*sin(11*\x r)});

\draw[domain=-7:-1.5,samples=120,smooth]
plot (\x,{14.8 + 0.32*sin(5.5*\x r + 0.8) + 0.15*sin(10*\x r)});

\draw[domain=-7:-1.5,samples=120,smooth]
plot (\x,{13.7 + 0.30*sin(4.8*\x r - 0.5) + 0.18*sin(12*\x r)});

\node[font=\Huge] at (-4.25,12.85) {$\vdots$};

\draw[domain=-7:-1.5,samples=120,smooth]
plot (\x,{11.9 + 0.35*sin(5.2*\x r + 0.3) + 0.16*sin(9*\x r)});

\draw[line width=0.9pt,-{Stealth[length=3mm,width=2mm]}] (-7,10.9) -- (-1.5,10.9);
\node[font=\large] at (-4.25,10.35) {Time};

\node[scale=1.35,font={\Huge\bfseries\sffamily}] at (-4.25,21)
{\textit{\textbf{Multivariate}}};

\node[scale=1.35,font={\Huge\bfseries\sffamily}] at (-4.25,19.5)
{\textit{\textbf{Time Series}}};

\draw[fullarrow] (-0.5,14.2) -- (2.0,14.2);


\draw (3.2,18.5) rectangle (4.4,9.8);

\foreach \x in {3.6,4.0}
{
    \draw (\x,18.5) -- (\x,9.8);
}

\foreach \y in {17.75,17,16.25,15.5,14.75,14,13.25,12.5,11.75,11,10.25}
{
    \draw (3.2,\y) -- (4.4,\y);
}

\draw (4.4,18.3) rectangle (4.65,9.8);
\draw (4.65,18.1) rectangle (4.9,9.8);

\draw (4.4,18.5) -- (4.9,18.1);
\draw (3.2,9.8) -- (4.65,9.8);
\draw (4.4,9.8) -- (4.9,9.8);

\node[font=\large] at (2.75,14.1) {$N$};
\node[font=\large] at (3.85,9.15) {$T$};
\node[font=\large] at (5,18.5) {$C$};

\node[scale=1.25,font={\Huge\bfseries\sffamily}] at (4.1,7.7)
{\textit{\textbf{Input Representation}}};

\node[scale=1.25,font={\Huge\bfseries\sffamily}] at (4.1,6.25)
{\textit{\textbf{$N \times C \times T$}}};


\draw (4.9,18.1) -- (10.7,16.8);
\draw (4.9,9.8) -- (10.7,11.6);

\draw (10.7,16.8) rectangle (11.7,11.6);
\draw (11.7,16.5) rectangle (12.1,11.9);
\draw (12.1,16.2) rectangle (12.5,12.2);

\node[scale=1.25,font={\Huge\bfseries\sffamily}] at (11.4,21)
{\textit{\textbf{Random 1-D}}};

\node[scale=1.25,font={\Huge\bfseries\sffamily}] at (11.4,19.5)
{\textit{\textbf{Convolution kernels}}};

\node[scale=1.25,font={\Huge\bfseries\sffamily}] at (11.4,18)
{\textit{\textbf{$N_{ker} \times \ell_{ker} $}}};

\draw[fullarrow] (14.0,14.2) -- (16.5,14.2);


\draw[rounded corners=5pt,line width=0.9pt] (17.0,17.4) rectangle (22.2,11.0);

\node[font=\normalsize] at (19.6,16.35) {$A_{i,\kappa,S_\kappa}(t)
=
\sum_{c \in S_\kappa}
\left(X_{i,c} * W_\kappa\right)(t)$};

\node[font=\normalsize] at (19.6,15.5) {$\forall i \in \{1,\dots,N\}$};

\node[font=\normalsize] at (19.6,15) {$\forall \kappa \in \{1,\dots,N_{ker}\}$};

\node[font=\normalsize] at (19.6,14.5) {$\forall t \in \{1,\dots,T-\ell_{ker}+1\}$};

\node[font=\normalsize] at (19.6,13.5) {$\mathcal{D} = \{d_1,\dots,d_C\}, 
\qquad
S_\kappa \subseteq \mathcal{D}$};

\node[font=\normalsize] at (19.6,13) {$M_\kappa \sim \mathcal{U}(1,\ldots,9)
$};
\node[font=\normalsize] at (19.6,12.5) {$S_\kappa \sim \mathcal{U}\left(\mathcal{P}_{M_\kappa}(\mathcal{D})\right)$};

\node[font=\normalsize] at (19.6,11.9) {$ \mathcal{P}_{M_\kappa}(\mathcal{D}) = \{\mathcal{H} \subseteq  \mathcal{D} : |\mathcal{H}| = M_\kappa\}$};

\node[scale=1.25,font={\Huge\bfseries\sffamily}] at (19.6,7.7)
{\textit{\textbf{Activation Map Aggregation}}};

\node[scale=1.25,font={\Huge\bfseries\sffamily}] at (19.6,6.25)
{\textit{\textbf{$N \times N_{ker} \times (T-\ell_{ker}+1)$}}};

\draw[fullarrow] (22.7,14.2) -- (25.2,14.2);


\foreach \i in {0,...,8}
{
    \draw
    ({26.2 + 0.35*\i},{17.0 - 0.35*\i}) --
    ({28.2 + 0.35*\i},{17.0 - 0.35*\i}) --
    ({28.2 + 0.35*\i},{15.0 - 0.35*\i}) --
    ({26.2 + 0.35*\i},{15.0 - 0.35*\i}) -- cycle;
}

\node[scale=1.25,font={\Huge\bfseries\sffamily}] at (29.0,21)
{\textit{\textbf{Pooling Operators}}};

\node[scale=1.25,font={\Huge\bfseries\sffamily}] at (29.0,19.5)
{\textit{\textbf{(PPV, ZC, ...)}}};

\node[scale=1.25,font={\Huge\bfseries\sffamily}] at (29.0,18)
{\textit{\textbf{$N \times N_{ker} \times N_{PO}$}}};

\draw[fullarrow] (32.0,14.2) -- (34.5,14.2);


\draw[line width=0.9pt] (36.0,15.3) rectangle (38,13.3);

\node[scale=1.25,font={\Huge\bfseries\sffamily}] at (37,9.7)
{\textit{\textbf{Feature}}};

\node[scale=1.25,font={\Huge\bfseries\sffamily}] at (37,8.25)
{\textit{\textbf{Selection}}};

\node[scale=1.25,font={\Huge\bfseries\sffamily}] at (37,6.75)
{\textit{\textbf{$N \times N_{ker}$}}};
\draw[fullarrow] (39.0,14.2) -- (41.5,14.2);


\draw[rounded corners=7pt,line width=0.9pt] (43.2,17.2) rectangle node {\Huge Ridge} (46.3,11.2);

\node[scale=1.25,font={\Huge\bfseries\sffamily}] at (44.75,19)
{\textit{\textbf{Training}}};

\end{circuitikz}
    }%
    \caption{Multivariate SelF-Rocket Training Architecture}
    \label{fig:Architecture}
 \end{figure*}
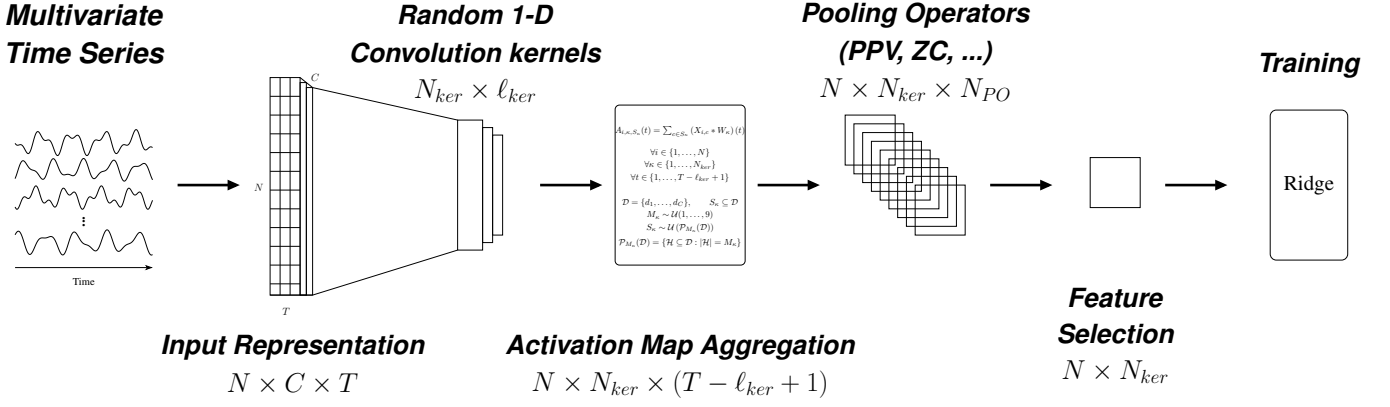

Our proposed approach, \emph{SelF-Rocket}, extends MiniRocket by automatically selecting the optimal combination of input representation (IR) and pooling operator (PO) through a feature selection module that retains the most discriminative features during training. SelF-Rocket searches over a set of IR-PO candidates. The input representations are the raw series, its  first-order difference (DIFF), and their concatenation (MIX). Besides the Proportion of Positive Values (PPV) used by MiniRocket, the candidate pooling operators include the Mean of Positive Values (MPV), the Mean of Indices of Positive Values (MIPV), the Longest Stretch of Positive Values (LSPV), and Zero Crossing (ZC), the latter counting the number of sign changes in the activation map~\cite{sapsanis2013improving}. Figure~\ref{fig:Architecture} illustrates the proposed multivariate architecture of \emph{SelF-Rocket}, which is built upon the univariate framework introduced in~\cite{lo2026timeseriesclassificationrandom}. The main modification concerns the activation map aggregation stage. For each kernel $\kappa$, a random subset of input dimensions $S_\kappa$ (with a maximum cardinality of 9) is selected, and the convolution responses computed over the channels in $S_\kappa$ are summed to produce a single activation map. Consequently, the computational complexity no longer scales linearly with the number of input dimensions, making the proposed method suitable for high-dimensional multivariate time series.
 \subsection{Proposed Pipeline}
\begin{figure}
    \centering
    \includegraphics[width=1\linewidth]{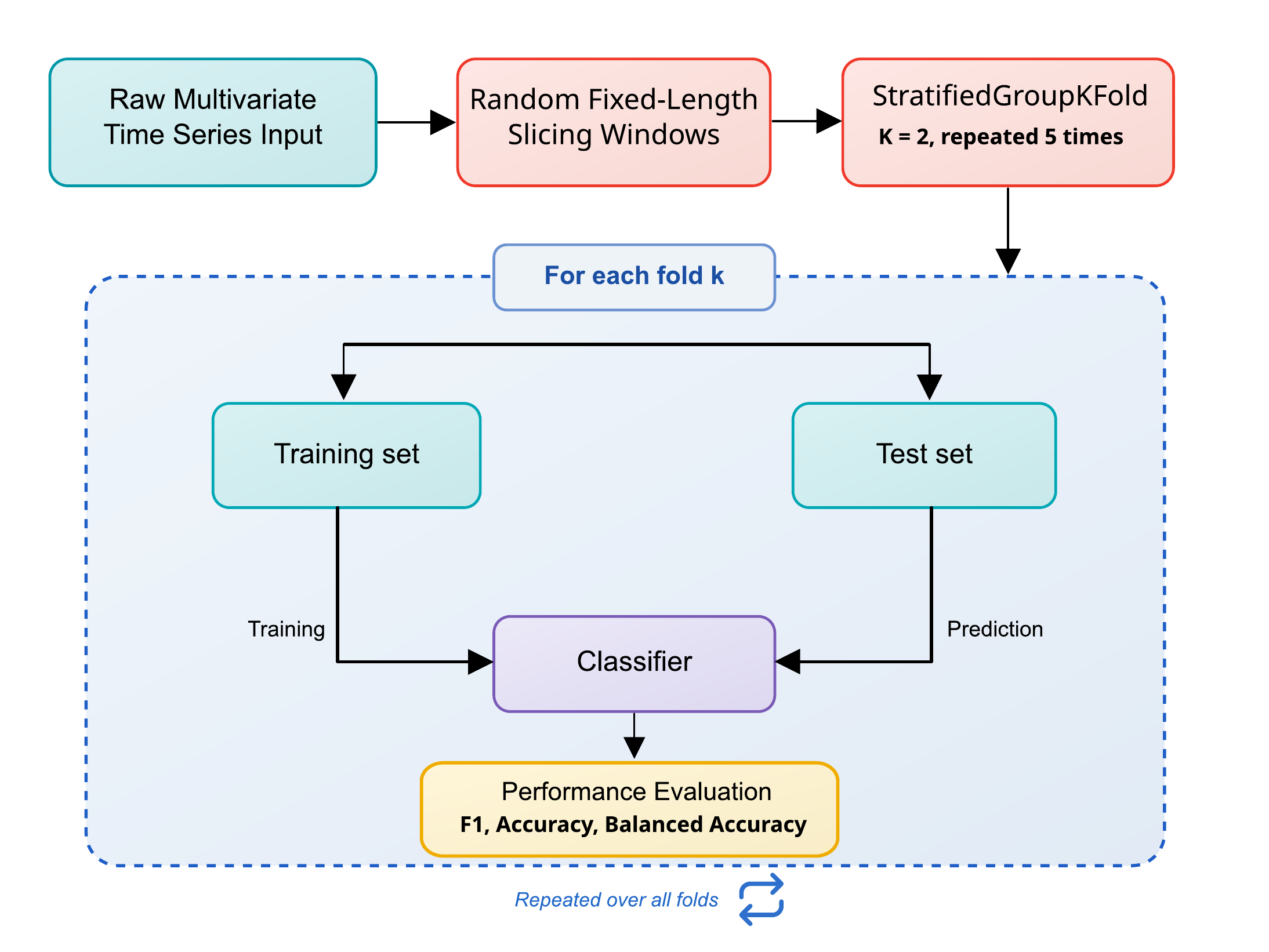}
    \caption{Proposed pipeline for multi-class fault classification using multivariate time series sensor data from a rotating machine}
    \label{fig:workflow}
\end{figure}

Most fault diagnosis methods for rotating machinery rely on feature engineering, whereby informative statistical features are extracted from measured signals prior to classification. These features are typically computed in the time, frequency, and time-frequency domains. Common time-domain features include the mean, median, root mean square (RMS), standard deviation, variance, skewness, kurtosis, and entropy. Frequency-domain features are generally derived from the signal spectrum, such as the amplitudes of the fundamental frequency and its sideband components, whereas time-frequency features are obtained by applying a wavelet transform followed by the extraction of statistical descriptors from the resulting coefficients.
The original recordings were randomly segmented into $N_w = 3$ fixed-length windows of size $w \in \{250,500,1000,2500,4500,5000\}$ samples. The number of extracted windows was intentionally limited to avoid excessive redundancy between highly correlated segments from the same recording, thereby promoting greater diversity in the training data. Besides improving robustness by reducing the dependence on positional information (i.e., signal phase), this preprocessing step produces equal-length time series, a prerequisite for convolution-based methods such as ROCKET. In contrast, traditional feature engineering approaches can operate on signals of varying lengths. To prevent data leakage caused by highly similar windows extracted from the same recording, a StratifiedGroupKFold cross-validation strategy was adopted, ensuring that all windows from a given recording belong to the same fold while preserving the class distribution. An overview of the complete training pipeline is presented in Figure~\ref{fig:workflow}.
\section{Experimental Results and Discussion}\label{sec:results}
\subsection{Experimental Setup}
All experiments were performed on a Dell Precision 5480 equipped with an Intel Core i9-13900H vPro processor (2.6 GHz), 32 GB of RAM, and running Windows 11. Our approach was implemented in Python 3.11.4, while the ROCKET-based baseline methods were implemented using the AEON Python Toolkit~\cite{aeon24jmlr}. The corresponding hyperparameters are reported in Table~\ref{tab:hyperparameters}.
\begin{table}[ht]
\caption{Hyperparameter settings for each ROCKET-based classifier. Methods marked with \textsuperscript{*} are implemented in the AEON Python Toolkit.}
\label{tab:hyperparameters}
\centering
\resizebox{\columnwidth}{!}{
\begin{tabular}{ll}
\toprule
\textbf{Classifiers} & \textbf{Hyperparameter Setting} \\
\midrule
MiniRocket\textsuperscript{*} & \texttt{MiniRocketClassifier(n\_jobs=-1)} \\
MultiRocket\textsuperscript{*} & \texttt{MultiRocketClassifier(n\_jobs=-1)} \\
Hydra\textsuperscript{*} & \texttt{HydraClassifier(n\_jobs=-1)} \\
Hydra + MR\textsuperscript{*} & \texttt{MultiRocketHydraClassifier(n\_jobs=-1)} \\
SelF-Rocket & \texttt{SelFRocket(num\_kernels=5000)} \\
\bottomrule
\end{tabular}
}
\end{table}
Each dataset was evaluated using a repeated two-fold StratifiedGroupKFold cross-validation protocol with five repetitions, yielding a total of 10 train-test splits. A two-fold strategy was adopted to maintain balanced training and test sets while maximizing the amount of unseen data used to evaluate the model's generalization performance. All methods were trained and evaluated on identical data partitions to ensure a fair comparison. Performance is reported as the mean classification score, mean execution time, and their corresponding standard errors across the 10 folds, providing an estimate of both predictive performance and computational efficiency.
The performance of a multi-class classifier is summarized by a confusion matrix, from which, for each class $k$, the \emph{True Positives} ($TP_k$), \emph{False Positives} ($FP_k$), \emph{True Negatives} ($TN_k$), and \emph{False Negatives} ($FN_k$) are derived. Predictive performance is assessed using three complementary metrics: \emph{Accuracy}, \emph{Balanced Accuracy}, and the \emph{macro-averaged F1-score}, defined from the per-class precision $P_k = TP_k/(TP_k+FP_k)$ and recall $R_k = TP_k/(TP_k+FN_k)$ as:
\begin{equation}
\mathrm{Accuracy}=\frac{\sum_{k=1}^{K}TP_k}{N},
\end{equation}
\begin{equation}
\mathrm{Balanced\ Accuracy}=\frac{1}{K}\sum_{k=1}^{K}R_k,
\end{equation}
\begin{equation}
F1_{\mathrm{macro}}=\frac{1}{K}\sum_{k=1}^{K}\frac{2P_k\cdot R_k}{P_k+R_k},
\end{equation}
where $K$ denotes the number of classes and $N$ the total number of instances.
\subsection{MaFaulDa Dataset}
\subsubsection{Classification Performance}
We first compare the classification performance of \emph{SelF-Rocket} with that of the other ROCKET-based methods using window sizes $w \in \{500,1000,2500,5000\}$ in the univariate setting and $w \in \{250,500\}$ in the multivariate setting. For the latter, all eight sensor channels were jointly exploited, whereas the univariate setting relied solely on the axial underhang bearing accelerometer signal. The corresponding results are reported in Table~\ref{tab:mafaulda_seg3_uni_multi_perf} (Appendix~\ref{app:results}). Across all window sizes and evaluation metrics, \emph{SelF-Rocket} consistently achieves the highest average predictive performance. For this dataset, it automatically selects \texttt{ZC\_MIX}, i.e., the Zero Crossing operator applied to the concatenation of the raw signal and its first-order difference, a combination not employed by the competing
ROCKET-based methods, which may contribute to the observed performance gains. In contrast, the performance gap between \emph{MultiRocket} / \emph{Hydra + MultiRocket} and the best methods widens as the window size increases, unlike methods relying on a single pooling operator.
\begin{figure}[!t]
    \centering
    \begin{subfigure}{\linewidth}
        \centering
        \includegraphics[width=\linewidth]{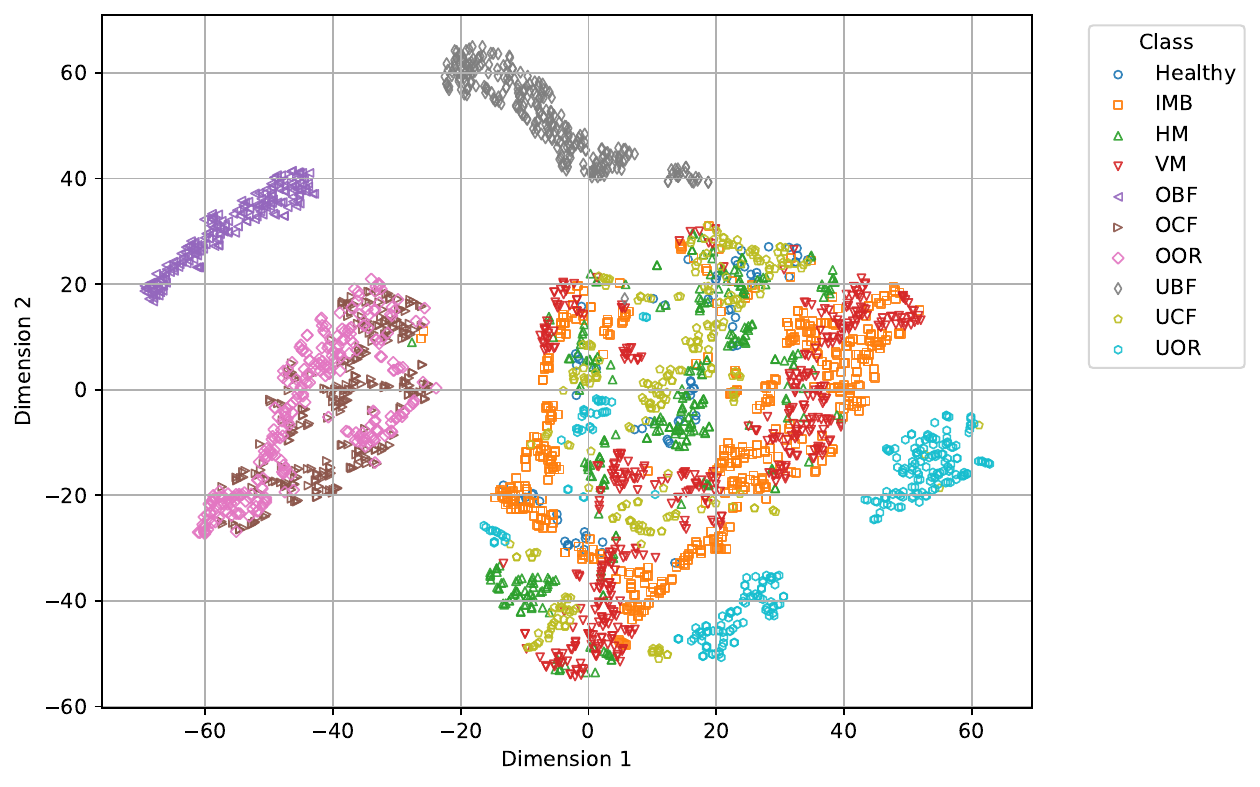}
        \caption{t-SNE $w = 250$}
    \end{subfigure}\\[4pt]
    \begin{subfigure}{\linewidth}
        \centering
        \includegraphics[width=\linewidth]{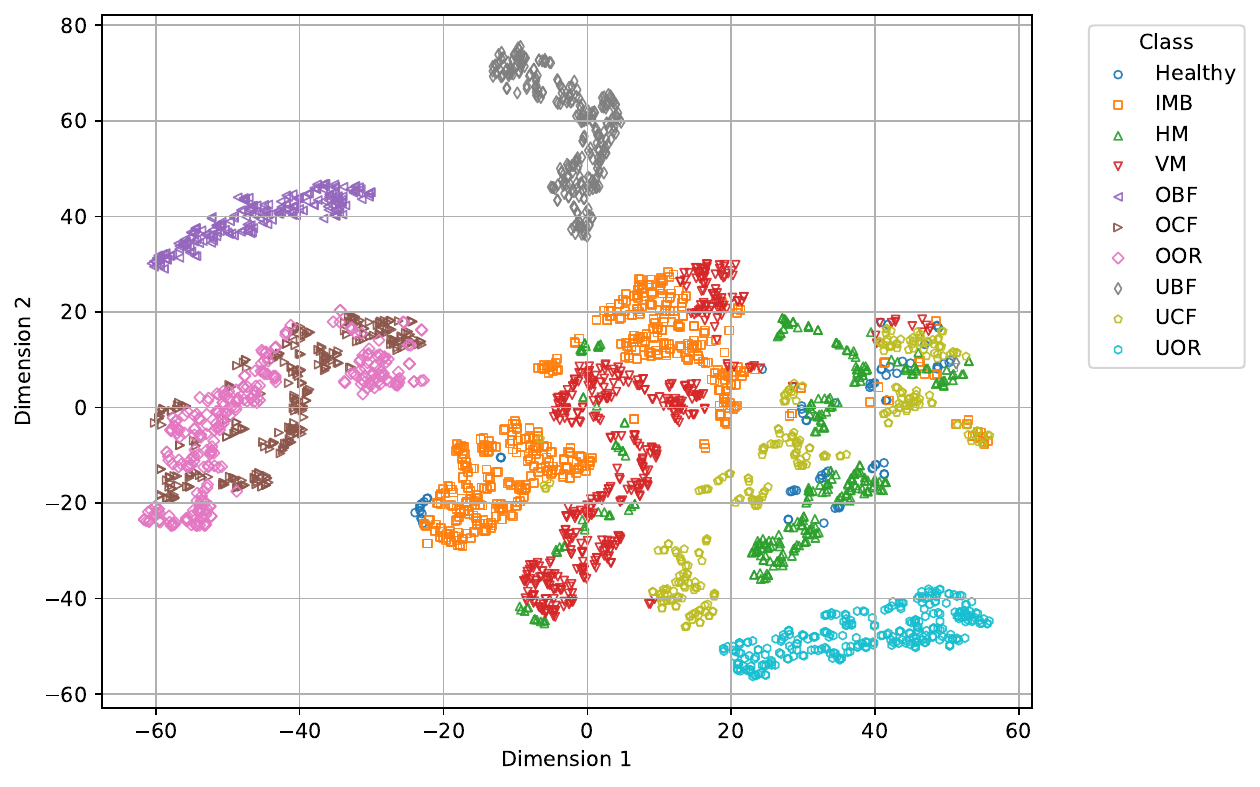}
        \caption{t-SNE $w = 500$}
    \end{subfigure}
    \caption{MaFaulDa multivariate t-SNE Projection of Feature Representations}
    \label{fig:TSNEMAFAULDA}
\end{figure}
To better understand these results, the representations of the two multivariate models ($w \in \{250,500\}$) were projected onto a two-dimensional space using t-distributed Stochastic Neighbor Embedding (t-SNE)~\cite{van2008visualizing}. The resulting embeddings, shown in Figure~\ref{fig:TSNEMAFAULDA}, reveal that $w = 500$ produces more compact and better-separated clusters than $w = 250$, suggesting more discriminative feature representations. Classes OBF, UBF, and UOR form clearly separated clusters, while OCF and OOR become partially distinguishable for $w = 500$. Conversely, IMB, HM, VM, UCF, and Healthy remain less separable, although the overlap between their representations is noticeably reduced when using larger windows, in agreement with the superior classification performance.
\subsubsection{Computational Performance}
Table~\ref{tab:mafaulda_seg3_uni_multi_time} (Appendix~\ref{app:results}) reports the computational performance of the evaluated methods. The reported training feature generation time includes random kernel generation, convolution, and feature extraction, while for \emph{SelF-Rocket} it additionally includes the selection of the optimal IR-PO combination. \emph{MiniRocket} achieves the shortest feature generation time for both training and inference. \emph{Hydra}, on the other hand, provides the lowest classifier training and inference times for $w \leq 2500$, owing to its smaller feature set ($512 \times 2 \times d$, where $d$ is the maximum dilation satisfying $2^{d} \leq \text{time series length}$).
\begin{figure}[!t]
    \centering
    \begin{subfigure}{\linewidth}
        \centering
        \includegraphics[width=\linewidth]{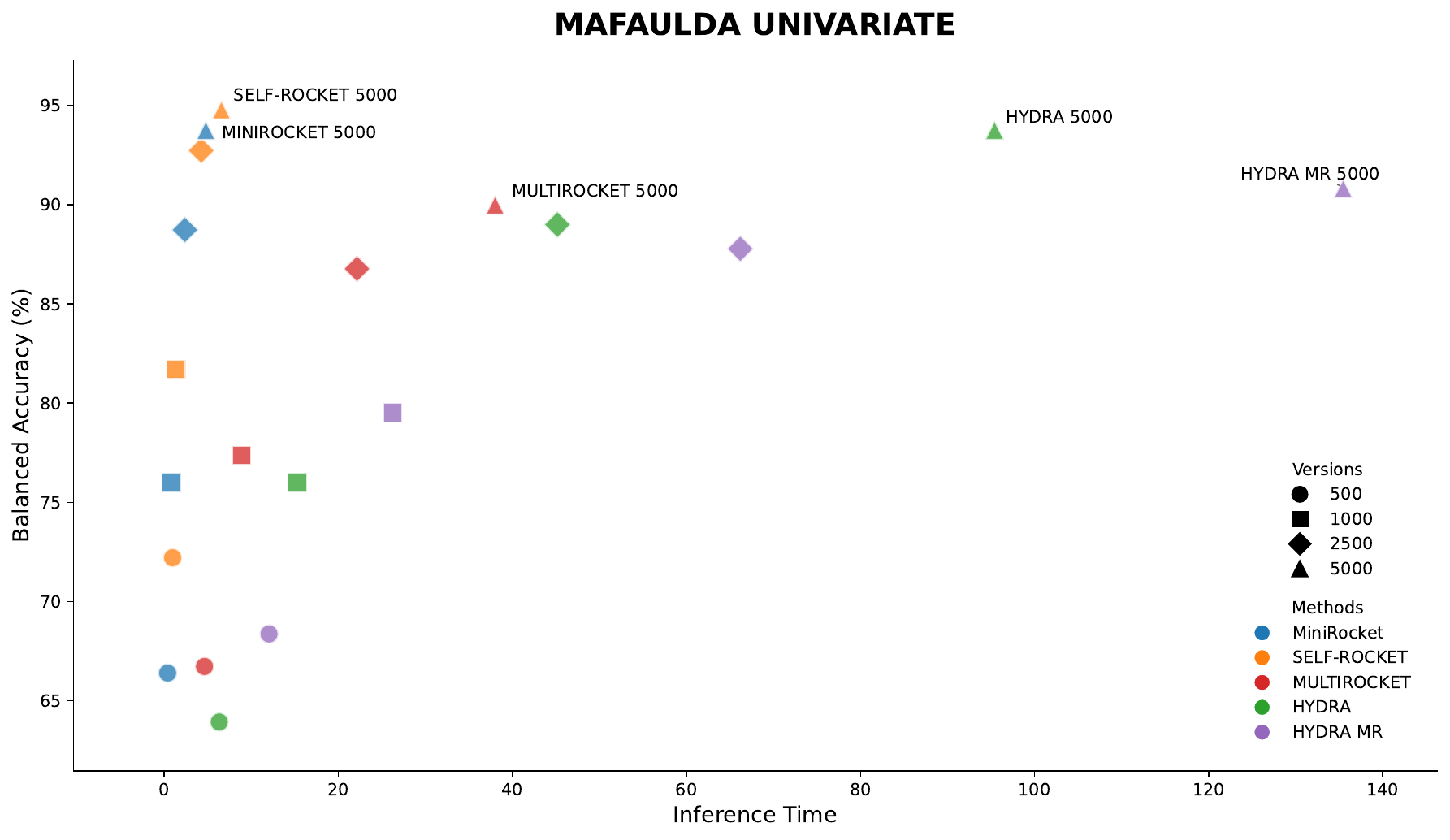}
        \caption{MaFaulDa univariate}
    \end{subfigure}\\[4pt]
    \begin{subfigure}{\linewidth}
        \centering
        \includegraphics[width=\linewidth]{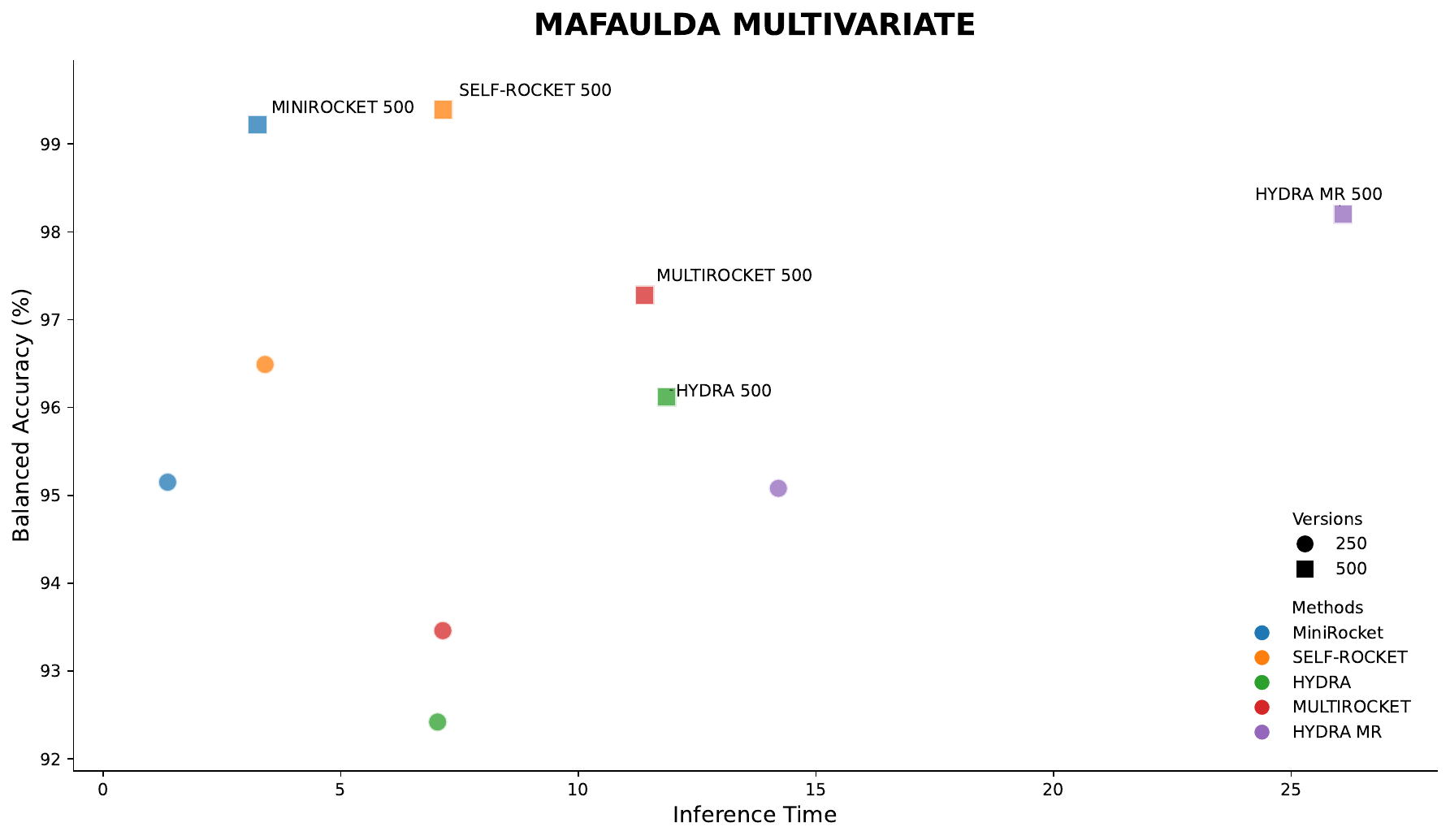}
        \caption{MaFaulDa multivariate}
    \end{subfigure}
    \caption{MaFaulDa univariate and multivariate Balanced Accuracy vs. Total Inference Time}
    \label{fig:perfsMAFAULDA}
\end{figure}
From a practical perspective, the most relevant computational metric is the total inference time, defined as the sum of feature generation and classifier inference times. Figure~\ref{fig:perfsMAFAULDA} compares all method--window size combinations in terms of total inference time and Balanced Accuracy. Although \emph{MiniRocket} offers the fastest feature extraction, \emph{SelF-Rocket} achieves the best overall accuracy--latency trade-off, combining the highest predictive performance with the second lowest total inference time. Consequently, it clearly outperforms competing approaches, including \emph{Hydra + MultiRocket}, which is widely regarded as one of the strongest ROCKET-based methods.
\subsection{ITSC-UDG Dataset}
\subsubsection{Classification Performance}
\begin{figure}[!t]
    \centering
    \begin{subfigure}{\linewidth}
        \centering
        \includegraphics[width=\linewidth]{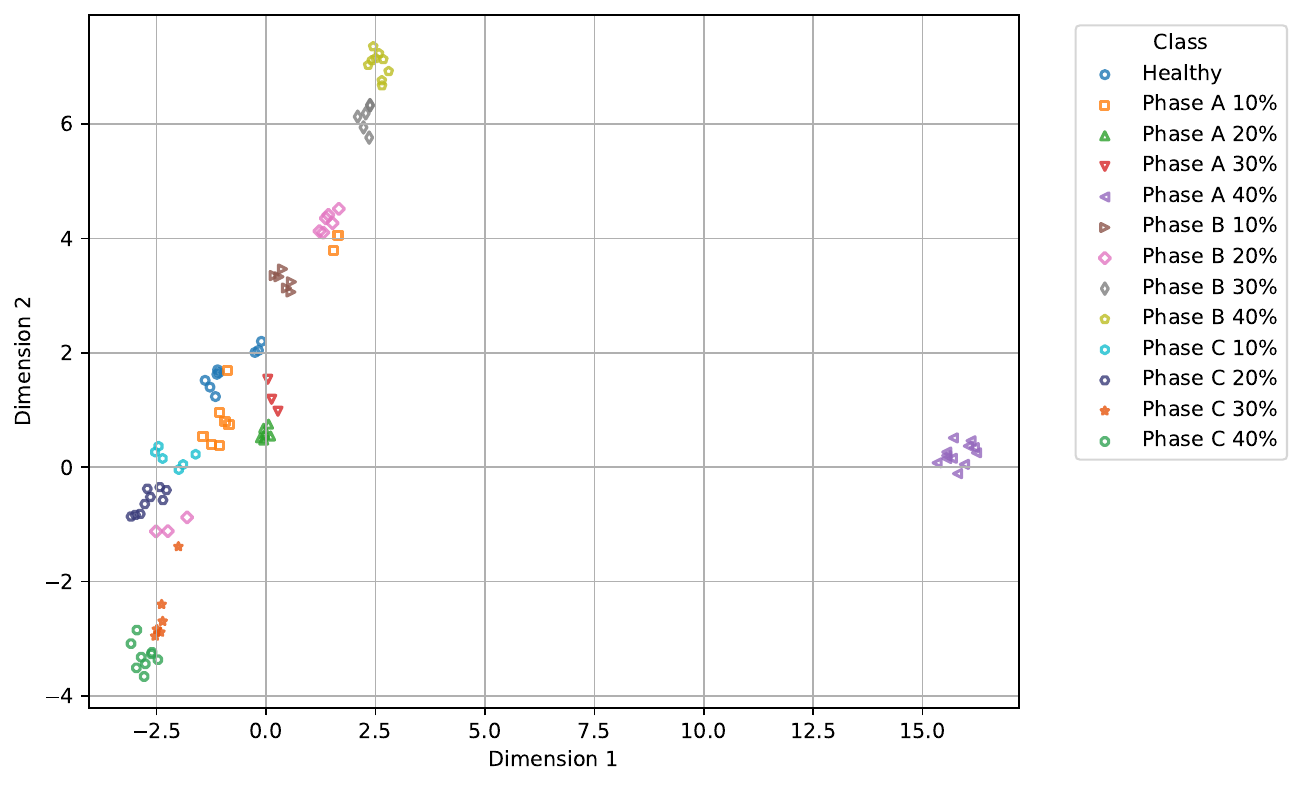}
        \caption{UMAP $w = 2500$}
    \end{subfigure}\\[4pt]
    \begin{subfigure}{\linewidth}
        \centering
        \includegraphics[width=\linewidth]{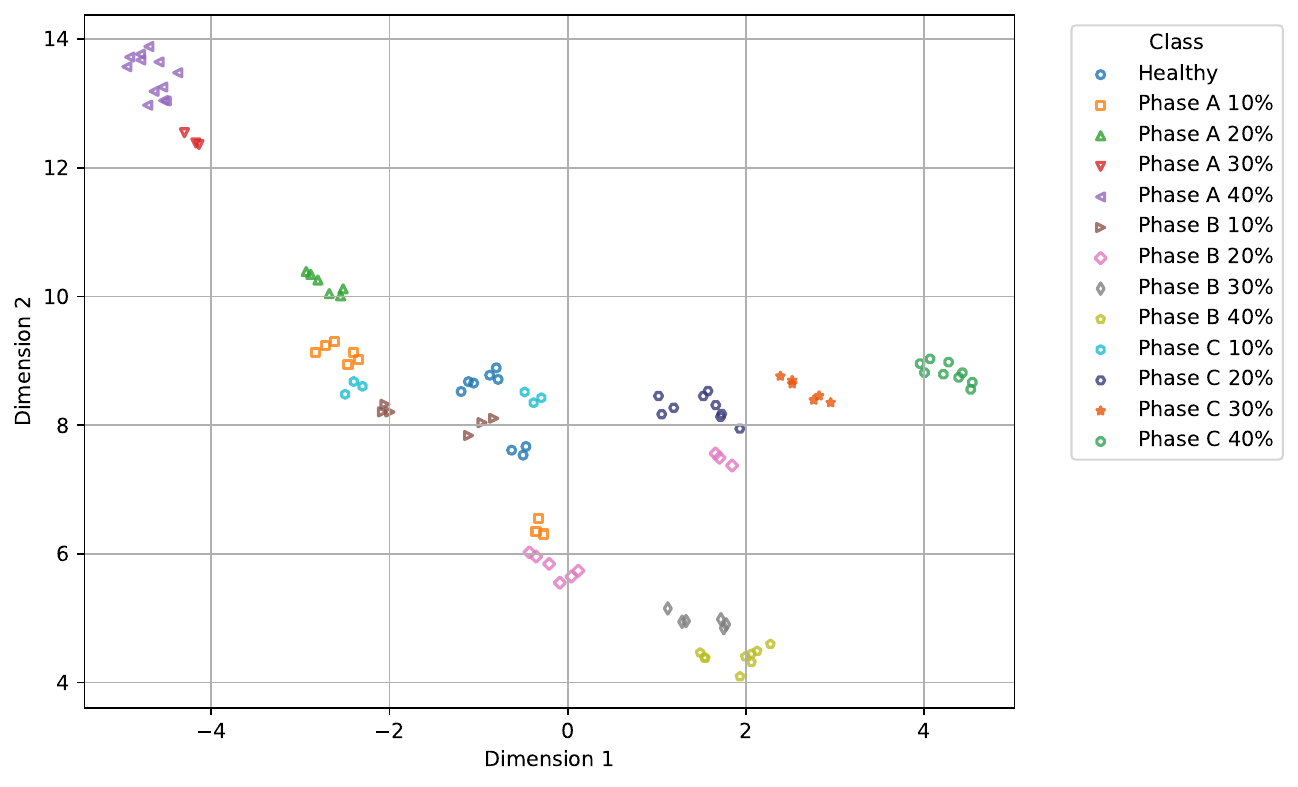}
        \caption{UMAP $w = 4500$}
    \end{subfigure}
    \caption{ITSC-UDG multivariate UMAP Projection of Feature Representations}
    \label{fig:UMAPITSC}
\end{figure}
\begin{figure}[!t]
    \centering
    \begin{subfigure}{\linewidth}
        \centering
        \includegraphics[width=\linewidth]{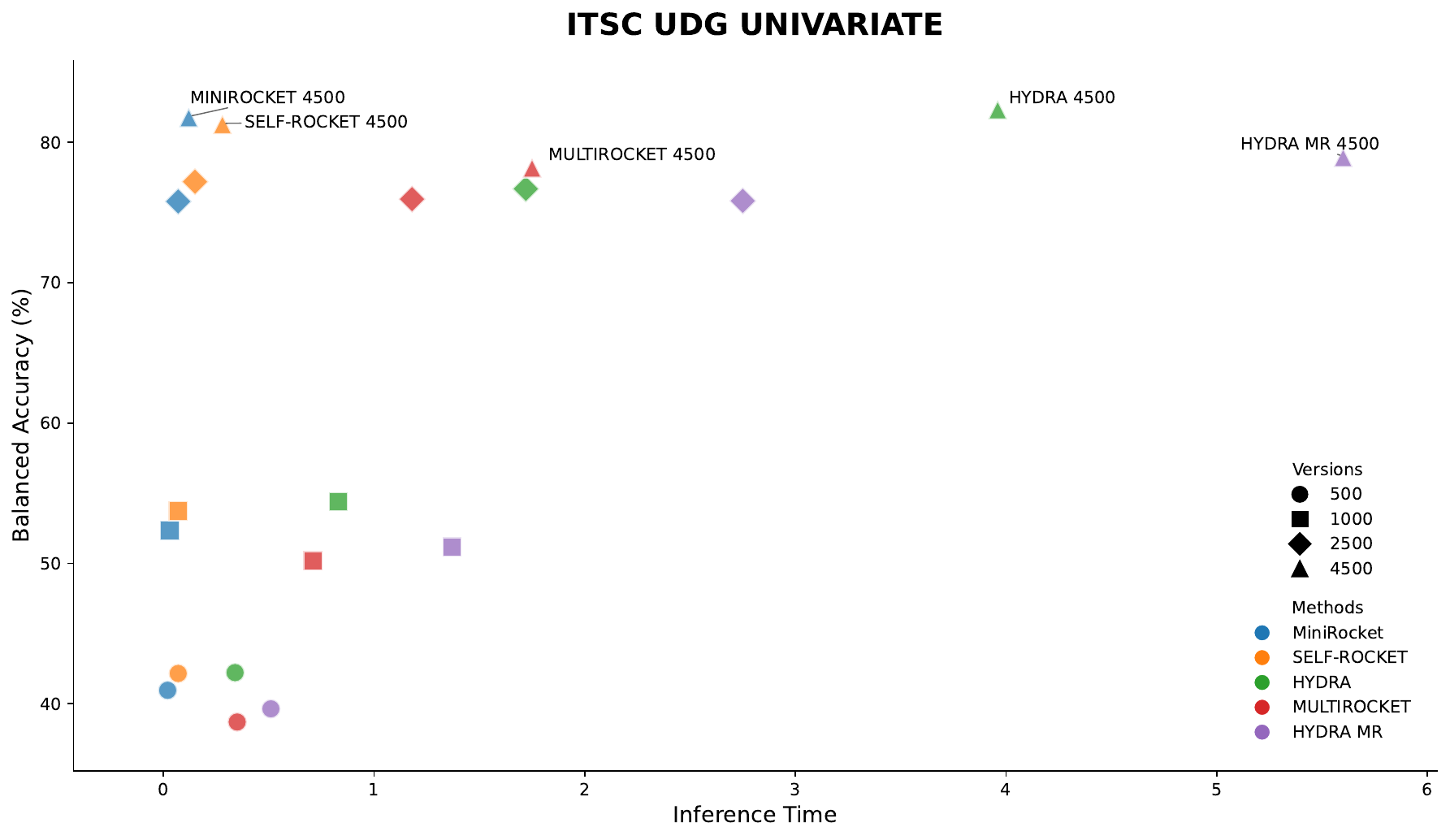}
        \caption{ITSC-UDG univariate}
    \end{subfigure}\\[4pt]
    \begin{subfigure}{\linewidth}
        \centering
        \includegraphics[width=\linewidth]{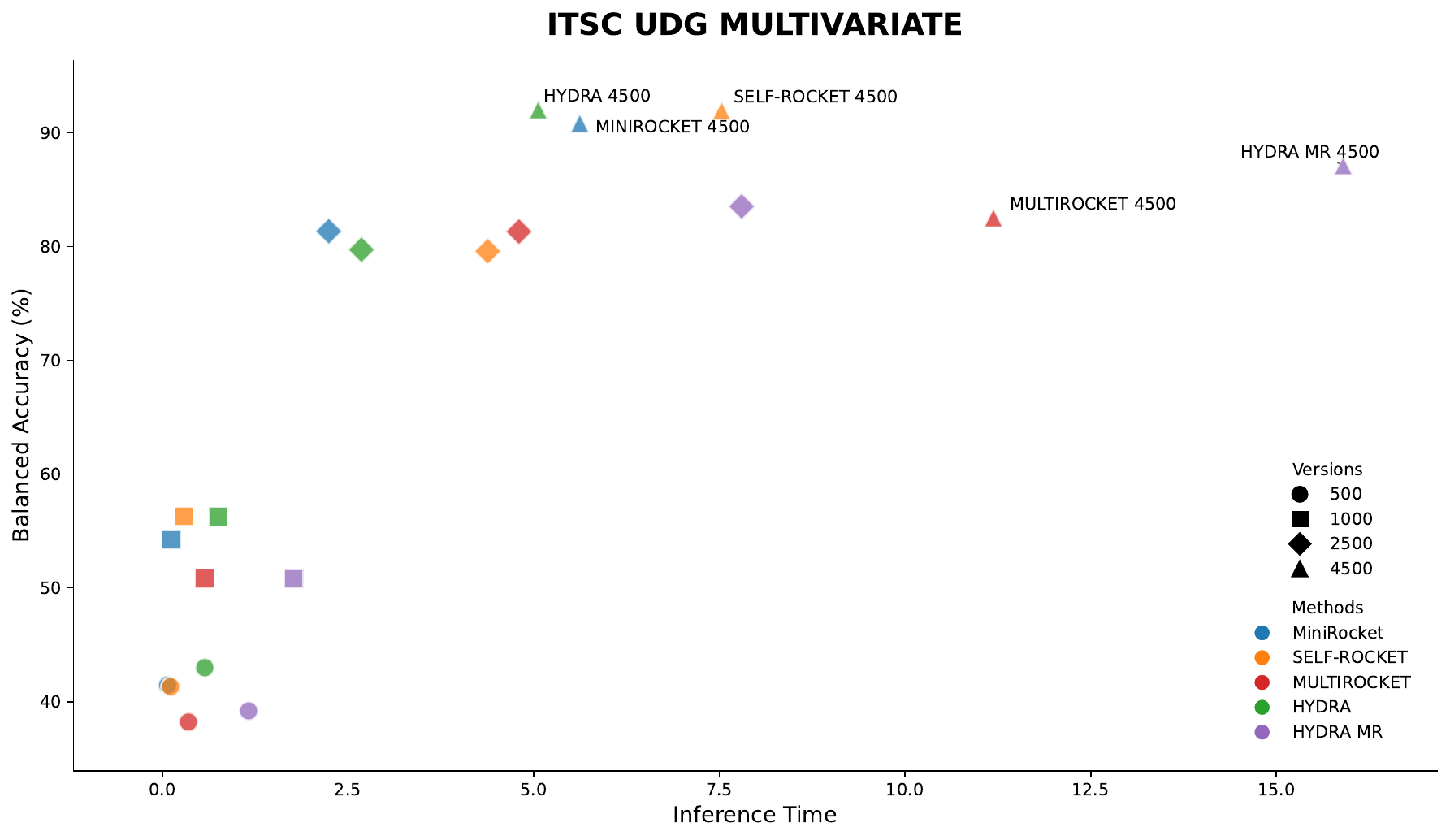}
        \caption{ITSC-UDG multivariate}
    \end{subfigure}
    \caption{ITSC-UDG univariate and multivariate Balanced Accuracy vs. Total Inference Time}
    \label{fig:perfsITSCUDG}
\end{figure}
For the ITSC-UDG dataset, we compared the univariate and multivariate classification performance of \emph{SelF-Rocket} with the other ROCKET-based methods using window sizes $w \in \{500,1000,2500,4500\}$. In the multivariate setting, all three stator current phase channels were jointly exploited, whereas the univariate setting relied solely on the phase-B current signal. Since the objective is to identify both the faulty phase and the fault severity, the multivariate representation is expected to be more informative, while the univariate experiments assess the discriminative power of a single phase measurement. The corresponding results are reported in Table~\ref{tab:perf_classification_uni_multi} (Appendix~\ref{app:results}).
Compared with MaFaulDa, the limited amount of available training data makes the selection of an optimal IR-PO combination considerably more challenging. Under the adopted evaluation protocol, approximately half of the 65 recordings are allocated to the test set in each fold, leaving relatively few training samples. Consequently, the performance differences between \emph{SelF-Rocket} and the competing methods are less pronounced than those observed on MaFaulDa, while the larger standard errors across folds indicate a higher variability of the estimated performance. Nevertheless, \emph{SelF-Rocket} remains highly competitive, consistently achieving results comparable to those of the best-performing ROCKET-based methods.
To further analyze the representations, Uniform Manifold Approximation and Projection (UMAP)~\cite{McInnes2018} was applied to the prediction embeddings produced by the two best-performing multivariate models ($w \in \{2500,4500\}$) with \emph{SelF-Rocket}. The resulting projections are shown in Figure~\ref{fig:UMAPITSC}. For $w = 2500$, the representations form relatively compact clusters, with the Phase-A 40\% class being particularly well separated. In contrast, the Healthy class lies in the densest region of the embedding, surrounded by neighboring classes, which agrees with the confusion matrices showing that most errors involve the Healthy condition and low-severity ITSC faults, which is expected, as low-severity faults produce weak signatures under no-load operation.
The projection obtained with $w = 4500$ reveals an even more structured organization of the feature space. The samples are progressively arranged along three distinct directions, each apparently associated with one of the three stator phases. Starting from the Healthy condition, the distance from the Healthy cluster increases with fault severity for the corresponding phase. Despite this improved organization, the Healthy neighborhood remains the densest region of the embedding, explaining why healthy conditions and low-severity faults remain the most difficult classes to discriminate.
\subsubsection{Computational Performance}
Finally, Table~\ref{tab:temps_execution_uni_multi} (Appendix~\ref{app:results}) and Figure~\ref{fig:perfsITSCUDG} summarize the computational performance and the corresponding accuracy--latency trade-offs for all evaluated window sizes $w$. In both the univariate and multivariate settings, \emph{SelF-Rocket} consistently achieves one of the best compromises between predictive performance and total inference time among the evaluated ROCKET-based methods.
\section{Conclusion}\label{sec:conclusion}
We introduced a multivariate extension of \emph{SelF-Rocket}, a random convolutional kernel method that automatically selects the most discriminative combination of input representation and pooling operator during training, and evaluated it for the multi-class diagnosis of mechanical and electrical faults. On the MaFaulDa and ITSC-UDG benchmarks, in both univariate and multivariate settings, it offers one of the best overall accuracy--latency trade-off among the evaluated ROCKET-based methods, achieving the highest performance on MaFaulDa while remaining highly competitive on the more challenging ITSC-UDG dataset. Its low inference latency suggests that it could be considered for embedded, real-time condition monitoring of electrical powertrains, as targeted in electric mobility applications.
Nevertheless, this study has several limitations. The evaluation was restricted to laboratory benchmark datasets acquired under controlled and steady-state operating conditions, and the results obtained on ITSC-UDG indicate that the automatic selection of the IR--PO combination becomes less reliable when only a limited number of training recordings are available. In addition, the proposed pipeline requires fixed-length input windows and labeled examples for each fault class. Future work will therefore focus on evaluating the robustness of the method under more challenging and realistic operating conditions, particularly under time-varying speed and load conditions, and on extending the evaluation to on-board deployment scenarios for real-time fault detection, in line with the objectives of the MYEL Joint Laboratory on the mobility and reliability of electrical powertrains.
\section*{Acknowledgment}
This research project, supported and financed by the French ANR (Agence Nationale pour la Recherche), is part of the Labcom (Laboratoire Commun) MYEL (MobilitY and Reliability of Electrical chain Lab) involving LSEE, LGI2A and CRITTM2A (ANR-22-LCV2-0001 MYEL).
\bibliographystyle{IEEEtran}
\bibliography{bibliography}
\appendices
\section{Detailed Performance Tables}\label{app:results}
This appendix gathers the detailed classification and computational performance tables for the MaFaulDa (Tables~\ref{tab:mafaulda_seg3_uni_multi_perf} and~\ref{tab:mafaulda_seg3_uni_multi_time}) and ITSC-UDG (Tables~\ref{tab:perf_classification_uni_multi} and~\ref{tab:temps_execution_uni_multi}) datasets.
\begin{table}[!ht]
\caption{Average classification performance (mean $\pm$ standard error) over 10 cross-validation folds on the MaFaulDa dataset for the univariate and multivariate settings. $w$ denotes the window size, and M. denotes the evaluation metric (Accuracy, Balanced Accuracy, and $F1_{\mathrm{macro}}$). MiniR, MultiR, H+MR, and SelFR denote MiniRocket, MultiRocket, Hydra+MultiRocket, and SelF-Rocket, respectively; the remaining column (Hydra) is shown under its full name.}
\label{tab:mafaulda_seg3_uni_multi_perf}
\centering
\begingroup
\scriptsize
\setlength{\tabcolsep}{1.6pt}
\renewcommand{\arraystretch}{1.08}
\resizebox{\columnwidth}{!}{%
\begin{tabular}{@{}llccccc@{}}
\toprule
$w$ & M. & MiniR & MultiR & Hydra & H+MR & SelFR \\
\midrule
\rowcolor{gray!12}
\multicolumn{7}{c}{\textbf{UNIVARIATE}} \\
\midrule
500  & Acc.  & $72.83{\pm}0.26$ & $72.23{\pm}0.21$ & $70.33{\pm}0.22$ & $74.01{\pm}0.20$ & $\bm{77.95{\pm}0.38}$ \\[1pt]
     & BAcc. & $66.40{\pm}0.22$ & $66.73{\pm}0.17$ & $63.93{\pm}0.19$ & $68.37{\pm}0.19$ & $\bm{72.21{\pm}0.41}$ \\[1pt]
     & F1    & $67.01{\pm}0.23$ & $67.18{\pm}0.18$ & $64.65{\pm}0.21$ & $68.92{\pm}0.21$ & $\bm{73.23{\pm}0.44}$ \\[2pt]
\cmidrule(lr){1-7}
1000 & Acc.  & $82.05{\pm}0.25$ & $82.66{\pm}0.15$ & $82.58{\pm}0.18$ & $84.72{\pm}0.18$ & $\bm{87.13{\pm}0.45}$ \\[1pt]
     & BAcc. & $76.00{\pm}0.37$ & $77.37{\pm}0.19$ & $76.00{\pm}0.19$ & $79.52{\pm}0.27$ & $\bm{81.70{\pm}0.53}$ \\[1pt]
     & F1    & $77.25{\pm}0.45$ & $78.48{\pm}0.20$ & $76.90{\pm}0.21$ & $80.71{\pm}0.30$ & $\bm{83.41{\pm}0.58}$ \\[2pt]
\cmidrule(lr){1-7}
2500 & Acc.  & $92.44{\pm}0.19$ & $91.04{\pm}0.15$ & $93.16{\pm}0.18$ & $92.02{\pm}0.14$ & $\bm{95.55{\pm}0.17}$ \\[1pt]
     & BAcc. & $88.73{\pm}0.37$ & $86.77{\pm}0.16$ & $89.00{\pm}0.23$ & $87.78{\pm}0.18$ & $\bm{92.73{\pm}0.24}$ \\[1pt]
     & F1    & $90.30{\pm}0.34$ & $88.21{\pm}0.18$ & $90.53{\pm}0.24$ & $89.19{\pm}0.19$ & $\bm{94.02{\pm}0.21}$ \\[2pt]
\cmidrule(lr){1-7}
5000 & Acc.  & $95.69{\pm}0.20$ & $93.37{\pm}0.15$ & $95.88{\pm}0.11$ & $94.00{\pm}0.17$ & $\bm{96.84{\pm}0.14}$ \\[1pt]
     & BAcc. & $93.76{\pm}0.38$ & $89.99{\pm}0.15$ & $93.76{\pm}0.19$ & $90.83{\pm}0.19$ & $\bm{94.79{\pm}0.32}$ \\[1pt]
     & F1    & $94.69{\pm}0.30$ & $91.37{\pm}0.14$ & $94.83{\pm}0.17$ & $92.12{\pm}0.20$ & $\bm{95.74{\pm}0.26}$ \\[2pt]
\midrule
\rowcolor{gray!12}
\multicolumn{7}{c}{\textbf{MULTIVARIATE}} \\
\midrule
250 & Acc.  & $97.37{\pm}0.09$ & $96.70{\pm}0.17$ & $96.42{\pm}0.05$ & $97.55{\pm}0.15$ & $\bm{98.34{\pm}0.08}$ \\[1pt]
    & BAcc. & $95.15{\pm}0.26$ & $93.46{\pm}0.39$ & $92.42{\pm}0.18$ & $95.08{\pm}0.39$ & $\bm{96.49{\pm}0.26}$ \\[1pt]
    & F1    & $96.10{\pm}0.20$ & $94.60{\pm}0.35$ & $93.58{\pm}0.20$ & $96.07{\pm}0.31$ & $\bm{97.33{\pm}0.20}$ \\[2pt]
\cmidrule(lr){1-7}
500 & Acc.  & $99.46{\pm}0.06$ & $98.47{\pm}0.09$ & $98.43{\pm}0.04$ & $99.03{\pm}0.09$ & $\bm{99.60{\pm}0.04}$ \\[1pt]
    & BAcc. & $99.22{\pm}0.08$ & $97.28{\pm}0.18$ & $96.12{\pm}0.19$ & $98.20{\pm}0.20$ & $\bm{99.39{\pm}0.07}$ \\[1pt]
    & F1    & $99.32{\pm}0.06$ & $97.87{\pm}0.14$ & $97.10{\pm}0.14$ & $98.59{\pm}0.15$ & $\bm{99.47{\pm}0.06}$ \\[2pt]
\bottomrule
\end{tabular}%
}
\endgroup
\end{table}
\begin{table}[!ht]
\caption{Average time performance (mean $\pm$ standard error) over 10 cross-validation folds on the MaFaulDa dataset for the univariate and multivariate settings. $w$ denotes the window size, and M. denotes the timing metric (TrFG = Training Feature Generation Time, ClTr = Classifier Training Time, TeFG = Test Feature Generation Time, and ClTe = Classifier Inference Time).}
\label{tab:mafaulda_seg3_uni_multi_time}
\centering
\begingroup
\scriptsize
\setlength{\tabcolsep}{1.3pt}
\renewcommand{\arraystretch}{1.08}
\resizebox{\columnwidth}{!}{%
\begin{tabular}{@{}llccccc@{}}
\toprule
$w$ & M. & MiniR & MultiR & Hydra & H+MR & SelFR \\
\midrule
\rowcolor{gray!12}
\multicolumn{7}{c}{\textbf{UNIVARIATE}} \\
\midrule
500  & TrFG & $\bm{0.48{\pm}0.01}$ & $4.11{\pm}0.17$ & $6.82{\pm}0.61$ & $12.21{\pm}0.86$ & $23.50{\pm}1.04$ \\[1pt]
     & ClTr & $4.51{\pm}0.12$ & $8.46{\pm}0.67$ & $\bm{4.32{\pm}0.20}$ & $9.31{\pm}0.55$ & $4.99{\pm}0.26$ \\[1pt]
     & TeFG & $\bm{0.38{\pm}0.00}$ & $4.61{\pm}0.45$ & $6.31{\pm}0.56$ & $12.03{\pm}0.93$ & $0.95{\pm}0.12$ \\[1pt]
     & ClTe & $0.09{\pm}0.00$ & $0.42{\pm}0.01$ & $\bm{0.05{\pm}0.00}$ & $0.48{\pm}0.01$ & $0.09{\pm}0.01$ \\[2pt]
\cmidrule(lr){1-7}
1000 & TrFG & $\bm{0.95{\pm}0.10}$ & $8.84{\pm}0.92$ & $14.28{\pm}0.77$ & $22.83{\pm}0.93$ & $27.23{\pm}0.97$ \\[1pt]
     & ClTr & $4.90{\pm}0.20$ & $7.88{\pm}0.30$ & $\bm{4.67{\pm}0.16}$ & $7.78{\pm}0.04$ & $4.73{\pm}0.28$ \\[1pt]
     & TeFG & $\bm{0.82{\pm}0.11}$ & $8.86{\pm}0.64$ & $15.27{\pm}0.95$ & $26.24{\pm}0.94$ & $1.34{\pm}0.28$ \\[1pt]
     & ClTe & $0.09{\pm}0.01$ & $0.42{\pm}0.01$ & $\bm{0.06{\pm}0.00}$ & $0.44{\pm}0.01$ & $0.07{\pm}0.01$ \\[2pt]
\cmidrule(lr){1-7}
2500 & TrFG & $\bm{2.59{\pm}0.25}$ & $17.25{\pm}0.35$ & $45.20{\pm}0.84$ & $66.05{\pm}1.02$ & $49.84{\pm}1.03$ \\[1pt]
     & ClTr & $4.90{\pm}0.30$ & $8.19{\pm}0.45$ & $\bm{4.74{\pm}0.17}$ & $10.30{\pm}0.52$ & $5.35{\pm}0.25$ \\[1pt]
     & TeFG & $\bm{2.36{\pm}0.26}$ & $22.14{\pm}0.80$ & $45.16{\pm}0.95$ & $66.19{\pm}1.01$ & $4.23{\pm}0.38$ \\[1pt]
     & ClTe & $0.10{\pm}0.01$ & $0.41{\pm}0.02$ & $\bm{0.07{\pm}0.00}$ & $0.50{\pm}0.01$ & $0.12{\pm}0.01$ \\[2pt]
\cmidrule(lr){1-7}
5000 & TrFG & $\bm{4.66{\pm}0.39}$ & $36.76{\pm}1.20$ & $96.27{\pm}0.89$ & $135.78{\pm}1.44$ & $77.01{\pm}1.98$ \\[1pt]
     & ClTr & $4.66{\pm}0.17$ & $9.09{\pm}0.39$ & $4.89{\pm}0.21$ & $8.51{\pm}0.26$ & $\bm{4.59{\pm}0.18}$ \\[1pt]
     & TeFG & $\bm{4.77{\pm}0.43}$ & $38.01{\pm}1.27$ & $95.40{\pm}1.08$ & $135.46{\pm}1.35$ & $6.56{\pm}1.14$ \\[1pt]
     & ClTe & $0.08{\pm}0.00$ & $0.41{\pm}0.02$ & $0.08{\pm}0.00$ & $0.50{\pm}0.01$ & $\bm{0.06{\pm}0.01}$ \\[2pt]
\midrule
\rowcolor{gray!12}
\multicolumn{7}{c}{\textbf{MULTIVARIATE}} \\
\midrule
250 & TrFG & $\bm{1.50{\pm}0.02}$ & $6.91{\pm}0.95$ & $7.30{\pm}0.53$ & $14.03{\pm}0.96$ & $23.89{\pm}1.12$ \\[1pt]
    & ClTr & $4.55{\pm}0.11$ & $8.33{\pm}0.47$ & $\bm{4.26{\pm}0.21}$ & $9.85{\pm}0.49$ & $4.92{\pm}0.33$ \\[1pt]
    & TeFG & $\bm{1.36{\pm}0.01}$ & $7.15{\pm}0.76$ & $7.04{\pm}0.50$ & $14.21{\pm}1.09$ & $3.41{\pm}0.65$ \\[1pt]
    & ClTe & $0.08{\pm}0.00$ & $0.43{\pm}0.02$ & $\bm{0.05{\pm}0.00}$ & $0.52{\pm}0.04$ & $0.08{\pm}0.01$ \\[2pt]
\cmidrule(lr){1-7}
500 & TrFG & $\bm{3.29{\pm}0.41}$ & $14.34{\pm}2.97$ & $11.21{\pm}1.00$ & $24.22{\pm}1.34$ & $27.60{\pm}0.70$ \\[1pt]
    & ClTr & $5.03{\pm}0.21$ & $8.95{\pm}0.51$ & $\bm{4.40{\pm}0.17}$ & $9.93{\pm}0.34$ & $4.47{\pm}0.11$ \\[1pt]
    & TeFG & $\bm{3.25{\pm}0.40}$ & $11.40{\pm}1.28$ & $11.86{\pm}0.70$ & $26.09{\pm}1.24$ & $7.16{\pm}0.80$ \\[1pt]
    & ClTe & $0.09{\pm}0.00$ & $0.43{\pm}0.02$ & $\bm{0.06{\pm}0.00}$ & $0.56{\pm}0.03$ & $0.08{\pm}0.00$ \\[2pt]
\bottomrule
\end{tabular}%
}
\endgroup
\end{table}
\begin{table}[!ht]
\caption{Average classification performance (mean $\pm$ standard error) over 10 cross-validation folds on the ITSC-UDG dataset for univariate and multivariate settings.}
\label{tab:perf_classification_uni_multi}
\centering
\begingroup
\scriptsize
\setlength{\tabcolsep}{1.3pt}
\renewcommand{\arraystretch}{1.08}
\resizebox{\columnwidth}{!}{%
\begin{tabular}{@{}llccccc@{}}
\toprule
$w$ & M. & MiniR & MultiR & Hydra & H+MR & SelFR \\
\midrule
\rowcolor{gray!12}
\multicolumn{7}{c}{\textbf{UNIVARIATE}} \\
\midrule
500  & Acc.  & $40.48{\pm}1.19$ & $37.91{\pm}1.31$ & $\mathbf{41.70{\pm}1.27}$ & $39.04{\pm}1.29$ & $41.28{\pm}1.53$ \\[1pt]
     & BAcc. & $40.96{\pm}1.25$ & $38.70{\pm}1.39$ & $\mathbf{42.22{\pm}1.14}$ & $39.64{\pm}1.42$ & $42.16{\pm}1.59$ \\[1pt]
     & F1    & $40.61{\pm}1.06$ & $38.31{\pm}1.12$ & $\mathbf{41.72{\pm}0.90}$ & $39.34{\pm}1.16$ & $41.54{\pm}1.21$ \\[2pt]
\cmidrule(lr){1-7}
1000 & Acc.  & $51.26{\pm}1.33$ & $48.69{\pm}1.95$ & $\mathbf{53.40{\pm}1.51}$ & $49.82{\pm}1.66$ & $52.68{\pm}2.01$ \\[1pt]
     & BAcc. & $52.35{\pm}1.67$ & $50.19{\pm}2.17$ & $\mathbf{54.40{\pm}1.84}$ & $51.18{\pm}1.89$ & $53.74{\pm}2.32$ \\[1pt]
     & F1    & $50.48{\pm}1.32$ & $48.63{\pm}2.04$ & $\mathbf{52.38{\pm}1.39}$ & $49.58{\pm}1.76$ & $51.80{\pm}1.88$ \\[2pt]
\cmidrule(lr){1-7}
2500 & Acc.  & $74.92{\pm}1.32$ & $73.88{\pm}2.43$ & $76.25{\pm}1.55$ & $73.98{\pm}2.34$ & $\mathbf{76.56{\pm}1.53}$ \\[1pt]
     & BAcc. & $75.79{\pm}1.36$ & $75.96{\pm}1.98$ & $76.69{\pm}1.75$ & $75.83{\pm}1.93$ & $\mathbf{77.20{\pm}1.74}$ \\[1pt]
     & F1    & $74.58{\pm}1.46$ & $73.27{\pm}2.29$ & $75.31{\pm}1.49$ & $73.35{\pm}2.24$ & $\mathbf{75.97{\pm}1.34}$ \\[2pt]
\cmidrule(lr){1-7}
4500 & Acc.  & $80.81{\pm}1.50$ & $76.08{\pm}1.88$ & $\mathbf{81.53{\pm}1.48}$ & $77.21{\pm}1.93$ & $80.51{\pm}1.97$ \\[1pt]
     & BAcc. & $81.77{\pm}1.33$ & $78.18{\pm}1.32$ & $\mathbf{82.33{\pm}1.16}$ & $78.93{\pm}1.35$ & $81.28{\pm}1.56$ \\[1pt]
     & F1    & $79.43{\pm}1.31$ & $74.23{\pm}1.70$ & $\mathbf{79.75{\pm}1.38}$ & $74.92{\pm}1.81$ & $79.00{\pm}1.64$ \\[2pt]
\midrule
\rowcolor{gray!12}
\multicolumn{7}{c}{\textbf{MULTIVARIATE}} \\
\midrule
500  & Acc.  & $40.88{\pm}1.55$ & $36.99{\pm}1.23$ & $\mathbf{42.61{\pm}1.50}$ & $38.24{\pm}1.00$ & $40.27{\pm}1.23$ \\[1pt]
     & BAcc. & $41.45{\pm}1.53$ & $38.21{\pm}1.35$ & $\mathbf{42.99{\pm}1.35}$ & $39.19{\pm}1.19$ & $41.32{\pm}1.34$ \\[1pt]
     & F1    & $40.92{\pm}1.35$ & $36.93{\pm}1.03$ & $\mathbf{42.50{\pm}1.12}$ & $38.35{\pm}0.97$ & $40.59{\pm}1.11$ \\[2pt]
\cmidrule(lr){1-7}
1000 & Acc.  & $53.30{\pm}1.39$ & $49.19{\pm}1.83$ & $\mathbf{55.65{\pm}1.43}$ & $49.62{\pm}1.76$ & $55.64{\pm}1.67$ \\[1pt]
     & BAcc. & $54.23{\pm}1.71$ & $50.83{\pm}1.99$ & $56.26{\pm}1.59$ & $50.79{\pm}1.99$ & $\mathbf{56.30{\pm}1.92}$ \\[1pt]
     & F1    & $52.66{\pm}1.24$ & $49.95{\pm}1.83$ & $54.83{\pm}1.35$ & $49.74{\pm}1.82$ & $\mathbf{55.05{\pm}1.45}$ \\[2pt]
\cmidrule(lr){1-7}
2500 & Acc.  & $80.67{\pm}1.16$ & $79.45{\pm}1.87$ & $79.14{\pm}1.01$ & $\mathbf{82.09{\pm}1.88}$ & $78.92{\pm}1.38$ \\[1pt]
     & BAcc. & $81.35{\pm}1.29$ & $81.32{\pm}1.49$ & $79.72{\pm}1.19$ & $\mathbf{83.53{\pm}1.54}$ & $79.59{\pm}1.56$ \\[1pt]
     & F1    & $79.98{\pm}0.97$ & $78.88{\pm}1.80$ & $78.51{\pm}0.97$ & $\mathbf{81.93{\pm}1.70}$ & $78.64{\pm}1.22$ \\[2pt]
\cmidrule(lr){1-7}
4500 & Acc.  & $90.14{\pm}1.05$ & $79.67{\pm}1.82$ & $\mathbf{91.45{\pm}1.25}$ & $85.00{\pm}1.15$ & $91.36{\pm}1.16$ \\[1pt]
     & BAcc. & $90.85{\pm}1.01$ & $82.54{\pm}1.37$ & $\mathbf{92.03{\pm}1.12}$ & $87.12{\pm}0.98$ & $91.97{\pm}1.11$ \\[1pt]
     & F1    & $90.02{\pm}1.06$ & $79.05{\pm}1.91$ & $91.42{\pm}1.24$ & $84.51{\pm}1.17$ & $\mathbf{91.46{\pm}1.13}$ \\[2pt]
\bottomrule
\end{tabular}%
}
\endgroup
\end{table}
\begin{table}[!ht]
\caption{Average time performance (mean $\pm$ standard error) over 10 cross-validation folds on the ITSC-UDG dataset for univariate and multivariate settings.}
\label{tab:temps_execution_uni_multi}
\centering
\begingroup
\scriptsize
\setlength{\tabcolsep}{1.3pt}
\renewcommand{\arraystretch}{1.08}
\resizebox{\columnwidth}{!}{%
\begin{tabular}{@{}llccccc@{}}
\toprule
$w$ & M. & MiniR & MultiR & Hydra & H+MR & SelFR \\
\midrule
\rowcolor{gray!12}
\multicolumn{7}{c}{\textbf{UNIVARIATE}} \\
\midrule
500  & TrFG & $\mathbf{0.18{\pm}0.05}$ & $0.46{\pm}0.02$ & $0.33{\pm}0.02$ & $0.68{\pm}0.03$ & $2.42{\pm}0.05$ \\[1pt]
     & ClTr & $0.04{\pm}0.00$ & $0.15{\pm}0.01$ & $0.04{\pm}0.00$ & $0.15{\pm}0.01$ & $\mathbf{0.03{\pm}0.00}$ \\[1pt]
     & TeFG & $\mathbf{0.02{\pm}0.00}$ & $0.35{\pm}0.02$ & $0.34{\pm}0.01$ & $0.51{\pm}0.05$ & $0.07{\pm}0.02$ \\[1pt]
     & ClTe & $0.01{\pm}0.00$ & $0.03{\pm}0.00$ & $0.01{\pm}0.00$ & $0.02{\pm}0.00$ & $\mathbf{0.00{\pm}0.00}$ \\[2pt]
\cmidrule(lr){1-7}
1000 & TrFG & $\mathbf{0.20{\pm}0.01}$ & $0.84{\pm}0.04$ & $0.68{\pm}0.02$ & $1.65{\pm}0.05$ & $2.93{\pm}0.08$ \\[1pt]
     & ClTr & $0.04{\pm}0.00$ & $0.15{\pm}0.00$ & $0.05{\pm}0.00$ & $0.17{\pm}0.00$ & $\mathbf{0.03{\pm}0.00}$ \\[1pt]
     & TeFG & $\mathbf{0.03{\pm}0.00}$ & $0.71{\pm}0.02$ & $0.83{\pm}0.02$ & $1.37{\pm}0.03$ & $0.07{\pm}0.01$ \\[1pt]
     & ClTe & $0.01{\pm}0.00$ & $0.02{\pm}0.00$ & $0.01{\pm}0.00$ & $0.03{\pm}0.00$ & $\mathbf{0.00{\pm}0.00}$ \\[2pt]
\cmidrule(lr){1-7}
2500 & TrFG & $\mathbf{0.40{\pm}0.01}$ & $1.57{\pm}0.10$ & $1.70{\pm}0.11$ & $3.40{\pm}0.18$ & $3.54{\pm}0.17$ \\[1pt]
     & ClTr & $\mathbf{0.03{\pm}0.00}$ & $0.12{\pm}0.01$ & $0.05{\pm}0.00$ & $0.16{\pm}0.01$ & $\mathbf{0.03{\pm}0.00}$ \\[1pt]
     & TeFG & $\mathbf{0.07{\pm}0.00}$ & $1.18{\pm}0.12$ & $1.72{\pm}0.11$ & $2.75{\pm}0.18$ & $0.15{\pm}0.00$ \\[1pt]
     & ClTe & $0.01{\pm}0.00$ & $0.02{\pm}0.00$ & $0.01{\pm}0.00$ & $0.02{\pm}0.00$ & $\mathbf{0.00{\pm}0.00}$ \\[2pt]
\cmidrule(lr){1-7}
4500 & TrFG & $\mathbf{0.67{\pm}0.00}$ & $2.41{\pm}0.18$ & $3.82{\pm}0.19$ & $6.08{\pm}0.35$ & $5.11{\pm}0.25$ \\[1pt]
     & ClTr & $\mathbf{0.03{\pm}0.00}$ & $0.11{\pm}0.01$ & $0.06{\pm}0.00$ & $0.16{\pm}0.00$ & $0.04{\pm}0.00$ \\[1pt]
     & TeFG & $\mathbf{0.12{\pm}0.00}$ & $1.75{\pm}0.16$ & $3.96{\pm}0.17$ & $5.60{\pm}0.29$ & $0.28{\pm}0.02$ \\[1pt]
     & ClTe & $\mathbf{0.00{\pm}0.00}$ & $0.02{\pm}0.00$ & $0.01{\pm}0.00$ & $0.02{\pm}0.00$ & $0.01{\pm}0.00$ \\[2pt]
\midrule
\rowcolor{gray!12}
\multicolumn{7}{c}{\textbf{MULTIVARIATE}} \\
\midrule
500  & TrFG & $\mathbf{0.23{\pm}0.05}$ & $0.50{\pm}0.03$ & $0.51{\pm}0.02$ & $1.38{\pm}0.03$ & $2.52{\pm}0.05$ \\[1pt]
     & ClTr & $0.03{\pm}0.00$ & $0.10{\pm}0.01$ & $0.04{\pm}0.00$ & $0.19{\pm}0.00$ & $\mathbf{0.02{\pm}0.00}$ \\[1pt]
     & TeFG & $\mathbf{0.07{\pm}0.00}$ & $0.35{\pm}0.04$ & $0.57{\pm}0.04$ & $1.16{\pm}0.02$ & $0.11{\pm}0.01$ \\[1pt]
     & ClTe & $0.01{\pm}0.00$ & $0.02{\pm}0.00$ & $0.01{\pm}0.00$ & $0.03{\pm}0.00$ & $\mathbf{0.00{\pm}0.00}$ \\[2pt]
\cmidrule(lr){1-7}
1000 & TrFG & $\mathbf{0.29{\pm}0.00}$ & $0.81{\pm}0.04$ & $0.68{\pm}0.03$ & $1.95{\pm}0.11$ & $3.54{\pm}0.06$ \\[1pt]
     & ClTr & $\mathbf{0.04{\pm}0.00}$ & $0.10{\pm}0.01$ & $\mathbf{0.04{\pm}0.00}$ & $0.17{\pm}0.01$ & $\mathbf{0.04{\pm}0.00}$ \\[1pt]
     & TeFG & $\mathbf{0.12{\pm}0.00}$ & $0.57{\pm}0.08$ & $0.75{\pm}0.05$ & $1.77{\pm}0.11$ & $0.29{\pm}0.01$ \\[1pt]
     & ClTe & $\mathbf{0.01{\pm}0.00}$ & $0.02{\pm}0.00$ & $\mathbf{0.01{\pm}0.00}$ & $0.02{\pm}0.00$ & $\mathbf{0.01{\pm}0.00}$ \\[2pt]
\cmidrule(lr){1-7}
2500 & TrFG & $\mathbf{2.49{\pm}0.16}$ & $5.28{\pm}0.67$ & $2.63{\pm}0.17$ & $8.45{\pm}0.64$ & $6.79{\pm}0.45$ \\[1pt]
     & ClTr & $0.04{\pm}0.00$ & $0.11{\pm}0.01$ & $0.06{\pm}0.00$ & $0.18{\pm}0.01$ & $\mathbf{0.03{\pm}0.00}$ \\[1pt]
     & TeFG & $\mathbf{2.24{\pm}0.16}$ & $4.80{\pm}0.68$ & $2.68{\pm}0.17$ & $7.80{\pm}0.62$ & $4.38{\pm}1.06$ \\[1pt]
     & ClTe & $0.01{\pm}0.00$ & $0.02{\pm}0.00$ & $0.01{\pm}0.00$ & $0.03{\pm}0.00$ & $\mathbf{0.00{\pm}0.00}$ \\[2pt]
\cmidrule(lr){1-7}
4500 & TrFG & $6.10{\pm}0.33$ & $12.12{\pm}0.77$ & $\mathbf{5.23{\pm}0.20}$ & $17.29{\pm}0.74$ & $14.37{\pm}0.51$ \\[1pt]
     & ClTr & $0.04{\pm}0.00$ & $0.11{\pm}0.00$ & $0.07{\pm}0.00$ & $0.16{\pm}0.01$ & $\mathbf{0.03{\pm}0.00}$ \\[1pt]
     & TeFG & $5.62{\pm}0.37$ & $11.19{\pm}0.62$ & $\mathbf{5.06{\pm}0.25}$ & $15.90{\pm}0.86$ & $7.53{\pm}1.09$ \\[1pt]
     & ClTe & $0.01{\pm}0.00$ & $0.02{\pm}0.00$ & $0.01{\pm}0.00$ & $0.02{\pm}0.00$ & $\mathbf{0.00{\pm}0.00}$ \\[2pt]
\bottomrule
\end{tabular}%
}
\endgroup
\end{table}
\end{document}